\documentclass[10pt,twocolumn,letterpaper]{article}

\usepackage{cvpr}              % To produce the CAMERA-READY version
\usepackage[accsupp]{axessibility}  % Improves PDF readability for those with disabilities.

\usepackage{color}
\usepackage{array}
\usepackage{colortbl}
\newcommand{\pub}[1]{\color{gray}{\scriptsize{[{#1}]}}}
\usepackage{multirow}

\usepackage{bm}
\usepackage{graphicx}
\usepackage{amsmath}
\usepackage{makecell}
\usepackage{arydshln}
\usepackage{hhline}

\definecolor{cvprblue}{rgb}{0.21,0.49,0.74}
\usepackage[pagebackref,breaklinks,colorlinks,allcolors=cvprblue]{hyperref}

\def\paperID{4488} % *** Enter the Paper ID here
\def\confName{CVPR}
\def\confYear{2025}

\title{Revisiting Audio-Visual Segmentation with Vision-Centric Transformer}

\author{Shaofei Huang$^{1,2}$\footnotemark[1] \quad
Rui Ling$^{3}$ \quad
Tianrui Hui$^{1}$\footnotemark[2] \quad
Hongyu Li$^{4}$ \quad \\
Xu Zhou$^{5}$ \quad
Shifeng Zhang$^{5}$ \quad
Si Liu$^{4}$\footnotemark[2] \quad
Richang Hong$^{1}$ \quad
Meng Wang$^{1}$ \\
$^{1}$School of Computer Science and Information Engineering, Hefei University of Technology \quad \\
$^{2}$Institute of Information Engineering, Chinese Academy of Sciences \quad \\
$^{3}$School of Computer Science and Engineering, Beihang University \quad \\
$^{4}$School of Artificial Intelligence, Beihang University \quad $^{5}$Sangfor Technologies}

\begin{document}
\maketitle

\renewcommand{\thefootnote}{\fnsymbol{footnote}}
\footnotetext[1]{Visiting scholar at Hefei University of Technology.}
\footnotetext[2]{Corresponding authors.}

\begin{abstract}
Audio-Visual Segmentation (AVS) aims to segment sound-producing objects in video frames based on the associated audio signal.
Prevailing AVS methods typically adopt an audio-centric Transformer architecture, where object queries are derived from audio features.
However, audio-centric Transformers suffer from two limitations: perception ambiguity caused by the mixed nature of audio, and weakened dense prediction ability due to visual detail loss.
To address these limitations, we propose a new Vision-Centric Transformer (VCT) framework that leverages vision-derived queries to iteratively fetch corresponding audio and visual information, enabling queries to better distinguish between different sounding objects from mixed audio and accurately delineate their contours.
Additionally, we also introduce a Prototype Prompted Query Generation (PPQG) module within our VCT framework to generate vision-derived queries that are both semantically aware and visually rich through audio prototype prompting and pixel context grouping, facilitating audio-visual information aggregation.
Extensive experiments demonstrate that our VCT framework achieves new state-of-the-art performances on three subsets of the AVSBench dataset.
The code is available at \href{https://github.com/spyflying/VCT_AVS}{https://github.com/spyflying/VCT\_AVS}.
\end{abstract}    
\section{Introduction}
\label{sec:intro}

\begin{figure}[!t]
    \centering
    \includegraphics[width=\linewidth]{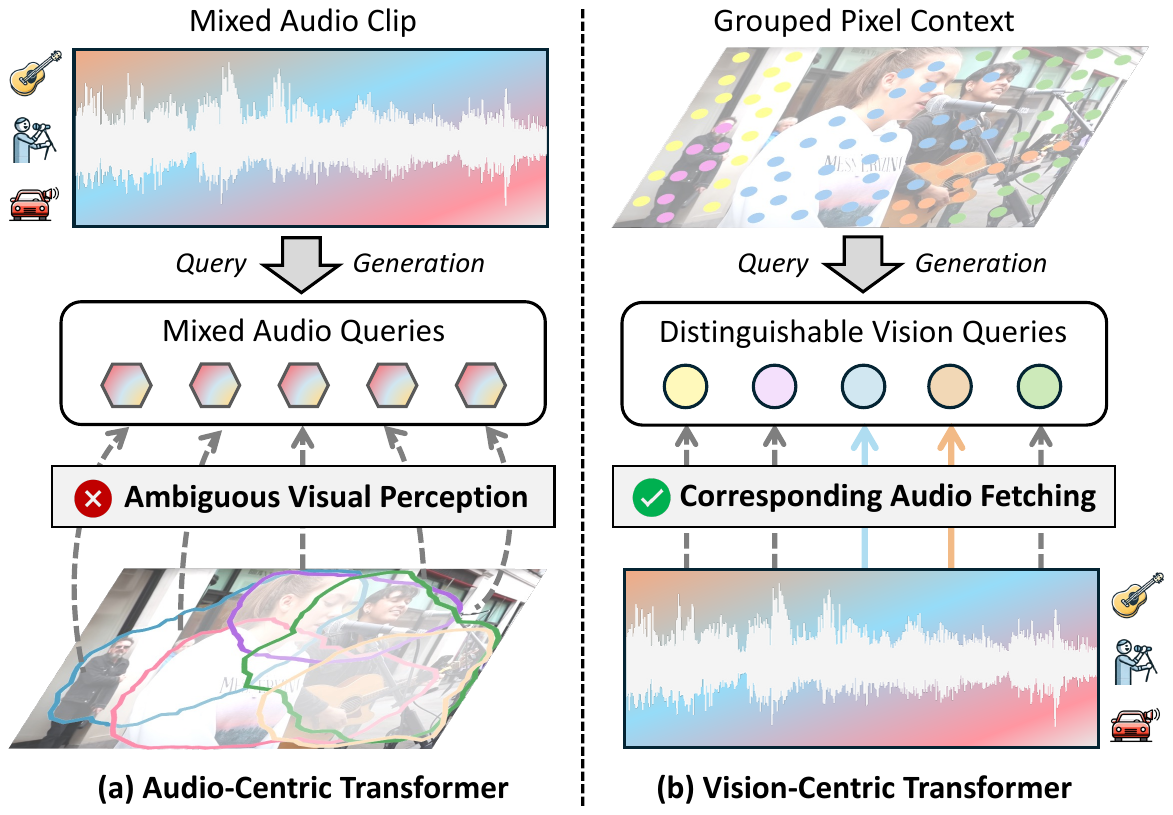}
    \caption{Motivation of our method. (a) Previous methods typically rely on audio-centric Transformers, where queries are derived from mixed sources of audio features, causing ambiguity in the sounding object perception process. (b) We propose the Vision-Centric Transformer, where queries are derived from visual features of different regions and interact with audio features to fetch corresponding audio information, achieving more accurate sound source distinguishment and object segmentation.}
    \label{fig:intro}
\end{figure}

Audio-Visual Segmentation (AVS)~\cite{TPAVI, TPAVI_J} is designed to identify and generate pixel-level segmentation masks of sound-emitting objects within visual frames by utilizing audio signals from the video.
This task requires models to determine regions associated with the sound source and to delineate the shape and boundaries of the sound-producing objects precisely.
Compared to Sound Source Localization (SSL)~\cite{senocak2018learning, hu2020discriminative, hu2022mix}, AVS places greater emphasis on fine-grained audio-visual understanding and pixel-level perception accuracy.
Based on these properties, the AVS task holds potential value in various application scenarios, including video understanding and editing, augmented and virtual reality, intelligent surveillance systems, \textit{etc}.

Prevailing AVS methods~\cite{GAVS, AVSegFormer, COMBO, CATR} typically employ an \textbf{audio-centric} Transformer architecture~\cite{DETR, Transformer}, where audio features are used as or integrated into object queries to identify the sounding objects in the video through layer-by-layer audio-visual interactions.
\textit{However, audio-centric Transformers present two main limitations.}
First, audio information is typically a mixture of various sound sources in real-world scenarios, including both semantically distinct sounds from in-screen sources and background noises from off-screen sources.
For example, the sample audio shown in Figure~\ref{fig:intro} contains mixed sounds from voice, guitar, and cars, with the last one not visible in the video frames.
This inherent mixture in audio causes potential confusion in perception for audio-derived queries, making them difficult to distinguish due to mutual interference.
Furthermore, noises from off-screen sources may introduce additional disturbance, causing audio-derived queries to yield false-positive predictions.
Second, AVS is essentially a vision-centric dense prediction task, where an object query must encompass both abstract audio semantics to identify the existence of the target sounding objects and concrete visual details to delineate their contours accurately.
However, audio-derived queries initially contain abstract audio semantics only.
The delayed integration of visual information may lead to the loss of concrete visual details, weakening the ability to make dense predictions.

To mitigate the aforementioned limitations of audio-centric Transformers, we centralize visual information in the query design and propose a new \textbf{Vision-Centric} Transformer (VCT) framework.
Within the VCT framework, object queries are initially derived from different image regions with abundant visual details, each serving as a candidate for a sounding object.
Through interactions with both audio and visual features across multiple Transformer decoder layers, these queries gradually become aware of their \textit{corresponding sound information} and \textit{fine-grained visual features}.
The former aids in determining whether an object is sound-emitting, while the latter helps in capturing the object's contours, thereby enabling comprehensive perception.
As shown in Figure~\ref{fig:intro}(b), these vision-derived queries, which focus on semantically distinct visual regions, can independently integrate their corresponding sound information from mixed audio, effectively excluding unrelated or noisy sound sources during interaction with audio features in the Transformer decoder layers.
By focusing on semantically relevant visual regions and sound information, vision-derived queries maintain superior discriminative ability, enabling them to better distinguish between different sounding objects in the scene and thereby alleviating the perceptual ambiguity caused by mixed audio.

Furthermore, to generate vision-derived queries that are both \textit{semantically aware} and \textit{visually rich}, we also introduce a novel Prototype Prompted Query Generation (PPQG) module within our VCT framework, thus laying a solid foundation for effective audio-visual information aggregation.
For the semantical aspect, we first define a set of prototypes for different categories of audio events.
Through contrastive learning with audio features, each prototype learns the corresponding acoustic semantic information from the audio signals.
The vision-derived queries are then prompted by these prototypes, making them aware of the potential audio events before interacting with the associated audio information.
This enables more targeted fetching of relevant audio information during subsequent interactions with the audio features.
For the visual aspect, since the queries are generated from visual features, they already contain a considerable amount of visual information.
What we need to do is enable vision-derived queries to initially focus on different regions, thus further enhancing their discriminative ability in the subsequent Transformer decoder.
To this end, we group pixel contexts into queries using an attention mechanism with a hard assignment, which encourages the perceptual focus of the queries to become more diverse.
With the PPQG module, our VCT framework can better harness the advantages of the vision-centric paradigm.

Contributions of our paper are summarized as follows:
(1) We propose a new Vision-Centric Transformer (VCT) framework that leverages vision-derived queries to fetch corresponding audio information and fine-grained visual features, thus realizing a comprehensive perception of sounding objects.
(2) We also introduce a novel Prototype Prompted Query Generation (PPQG) module to generate vision-derived queries that focus on different image regions with the awareness of occurring audio event categories, facilitating later information aggregation.
(3) Extensive experiments on three subsets of the AVSBench dataset demonstrate that our method outperforms previous state-of-the-art methods.
% \begin{itemize}
%     \item 
%     \item 
%     \item 
% \end{itemize}
\section{Related Works}
\label{sec:related_works}

\subsection{Sound Source Localization}
Sound Source Localization (SSL)~\cite{arandjelovic2017look, senocak2018learning, arandjelovic2018objects, hu2020discriminative, mo2022localizing, hu2022mix, kim2024learning} focuses on determining the location of sound-producing objects within a video by identifying visual areas that correspond to audio cues. This task aims to achieve a localized understanding of the scene at a patch level without emphasizing the exact shape of the sounding objects.
Hu~\textit{et al.}~\cite{hu2020discriminative} leverage audiovisual consistency as a self-supervised signal to match the category distribution differences between audio and visual inputs, effectively isolating sounding objects and filtering out silent ones.
Unsupervised SSL is explored in EZ-VSL~\cite{mo2022localizing} where a multiple-instance contrastive learning framework and an object-guided localization scheme are proposed to improve localization accuracy.
Kim \textit{et al.}~\cite{kim2024learning} achieve the localization of multiple sound sources without prior knowledge by iteratively identifying objects and applying object similarity-aware clustering loss.
Compared to SSL, AVS methods usually require more accurate visual representations to segment the sounding objects.

\begin{figure*}[!t]
    \centering
    \includegraphics[width=\linewidth]{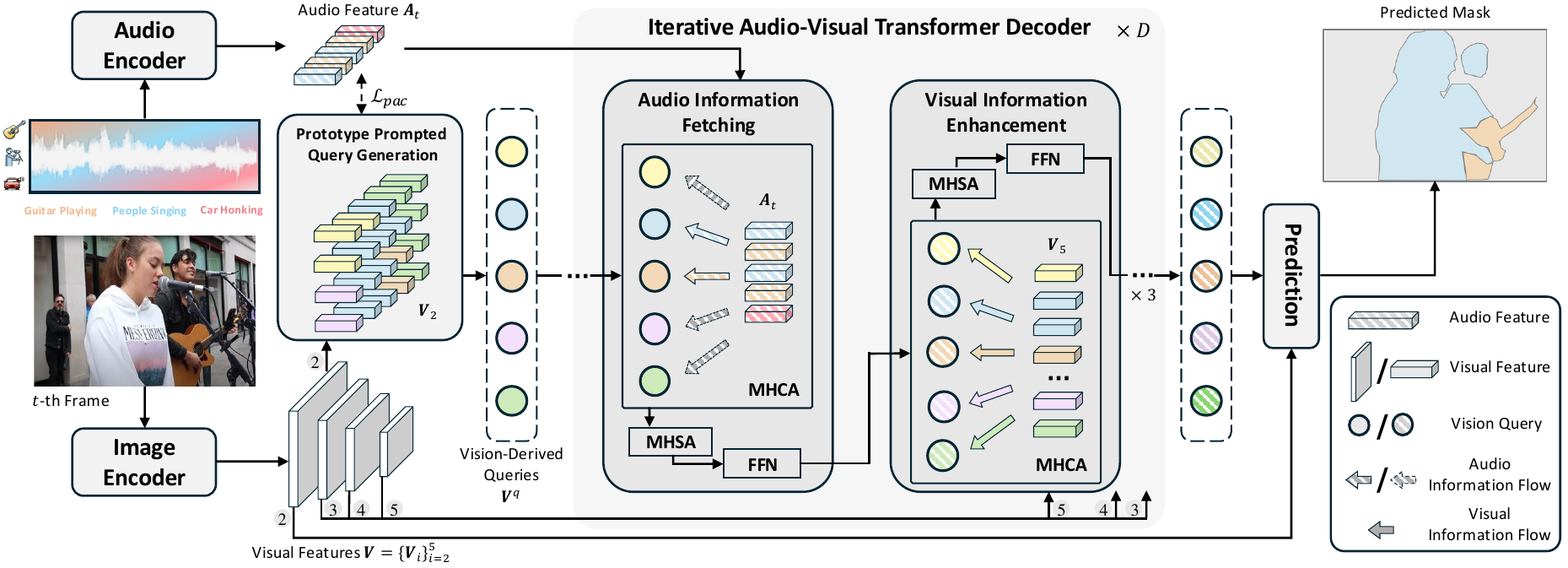}
    \caption{Overall architecture of our proposed VCT. The audio feature and multi-scale visual features are first extracted by the encoders. Taking the largest visual feature as input, our proposed Prototype Prompted Query Generation (PPQG) module generates vision-derived queries, which are then fed into the iterative audio-visual Transformer decoder. Through consecutive audio information fetching and visual information enhancement blocks, vision-derived queries are further refined to yield the final segmentation masks and categories.}
    \label{fig:pipeline}
\end{figure*}

\subsection{Audio-Visual Segmentation}
Different from sound source localization, Audio-Visual Segmentation (AVS)~\cite{TPAVI, TPAVI_J} aims to localize the sound-emitting objects with a more precise segmentation mask rather than a coarse heat map.
Some methods adopt the fusion-based architecture~\cite{TPAVI, TPAVI_J, ECMVAE, AVSBG} which first performs a multimodal fusion of features from the audio and visual encoders, then input the fused feature into a decoder that conducts per-pixel classification to generate the predicted mask.
For example, TPAVI~\cite{TPAVI, TPAVI_J} attentively aggregates the audio features with visual features from different stages of the encoder and performs an FPN-style fusion through the decoder.
Other Transformer-based methods~\cite{AQFormer, CATR, AVSC, BAVS, AVSegFormer, GAVS, COMBO, QDFormer, CPM, SelM, AL-Ref-SAM2} follow the mask classification paradigm of Mask2Former~\cite{Mask2Former} where queries are derived largely from audio information to discover sounding objects through Transformer decoder layers.
For example, COMBO~\cite{COMBO} enhances learnable queries by summing with audio features and performs bilateral fusion with visual features to construct more comprehensive queries.
We refer to the framework of these methods as audio-centric Transformers, which suffer from ambiguous perception and visual detail loss.
Apart from improvements in network architecture, recent works~\cite{UFE, TeSO, AVSBias} also focus on the intrinsic analysis of the AVS task and data.
In this paper, we propose a vision-centric Transformer framework to address the limitations of ambiguous perception and visual detail loss introduced by audio-centric Transformers.

\subsection{Query Designs for Transformers}
The three key components of the attention mechanism in Transformer~\cite{Transformer} models are query, key, and value, where the role of the query is to search for the most important information in the input sequence for the current position.
DETR~\cite{DETR} designs learnable object queries with image features as keys and values to detect object instances.
DINO~\cite{DINO} further introduces positional and content queries to more effectively capture both positional and categorical information for better identifying targets in complex scenes.
For multimodal tasks, ReferFormer~\cite{ReferFormer} and AQFormer~\cite{AQFormer} exploit language and audio as queries respectively to distinguish the referred objects by directly aggregating visual features relevant to the multimodal cues.
In this paper, we centralize visual information in the query design with a PPQG module to generate vision-derived queries that are both semantically aware and visually rich.
\section{Method}
\label{sec:method}

The overall framework of our proposed VCT is illustrated in Figure~\ref{fig:pipeline}.
Given an input video $\mathcal{V} \in \mathbb{R}^{T \times H \times W \times 3}$ with $T$ frames and an audio clip $\mathcal{A}$, AVS aims to produce the segmentation masks $\mathcal{Y} \in \mathbb{R}^{T \times H \times W \times K}$ of sounding objects, where $K$ is the number of audio event categories ($K = 1$ for category-agnostic settings, \textit{e.g.}, S4 and MS3).
We first feed the audio clip and multiple video frames into an audio encoder and an image encoder respectively to obtain the corresponding audio feature and multi-scale visual features.
The largest visual feature is then fed into our PPQG module to generate vision-derived queries, which interact with audio features and visual features iteratively in the Transformer decoder for further refinement.
Mask and category predictions are obtained based on refined vision-derived queries.

\subsection{Visual and Audio Feature Extraction}
\label{sec:method_feat_extraction}
For each input frame, we adopt a visual encoder (\textit{e.g.}, Swin Transformer~\cite{liu2021swin}) to extract multi-scale image features $\bm{V} = \{\bm{V}_i\}_{i=2}^5$, where $\bm{V}_i \in \mathbb{R}^{H_i \times W_i \times C_i^v}$, $H_i = \frac{H}{2^i}$, $W_i = \frac{W}{2^i}$, and $C_i^v$ are the height, width, and channel numbers of visual feature from the $i$-th stage of the visual encoder.
For the audio clip associated with the input video, we follow TPAVI~\cite{TPAVI} to utilize VGGish~\cite{VGGish} for audio feature extraction.
The original audio clip $\mathcal{A}$ is first converted to a mono output $\mathcal{A}_m \in \mathbb{R}^{L \times 96 \times 64}$ with a resampling rate of $16$ kHz to standardize the audio data, where $L$ varies according to the audio duration.
Then, we compute the time-frequency representation of $\mathcal{A}_m$ by applying Short-Time Fourier Transform to obtain its mel-spectrogram $\mathcal{A}_s \in \mathbb{R}^{T \times 96 \times 64}$, where $T$ is the number of frames.
$\mathcal{A}_s$ is further fed into the VGGish model with the final compression operation removed to extract the audio feature $\bm{A} \in \mathbb{R}^{T \times S \times C^a}$ where $S$ denotes the size of the mel-spectrogram feature, $C^a$ denotes its channel number.

\begin{figure}[!htbp]
    \centering
    \includegraphics[width=\linewidth]{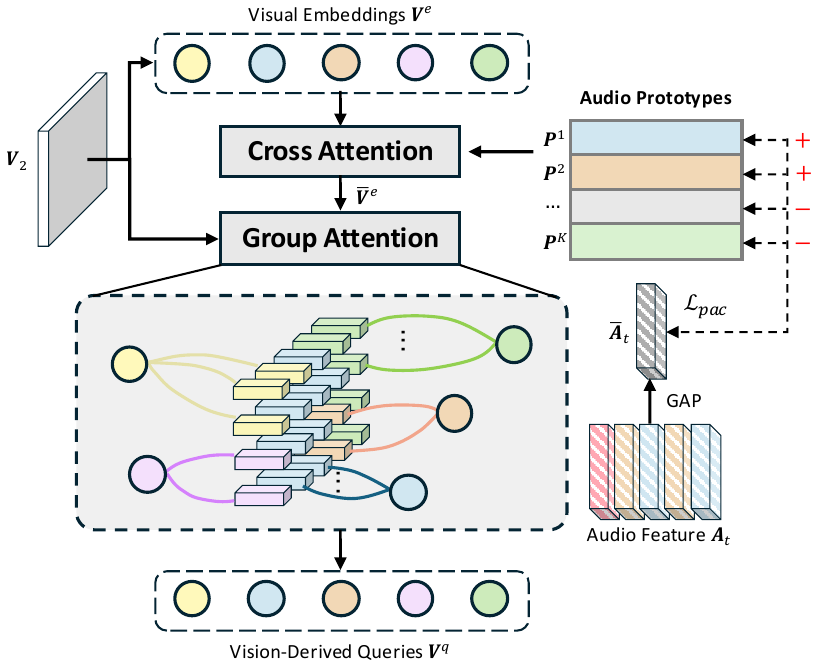}
    \caption{Details of Prototype Prompted Query Generation.}
    \label{fig:query}
\end{figure}

\subsection{Prototype Prompted Query Generation}
\label{sec:method_ppqg}

The goal of the Prototype Prompted Query Generation (PPQG) module is to generate vision-derived queries that contain both rich visual details and audio semantics.
As shown in Figure~\ref{fig:query}, the computation process of our PPQG module mainly consists of three steps.

\textbf{Visual embedding aggregation.}
We first perform spatial information aggregation on the high-resolution image feature $\bm{V}_2$ from the $2$-nd stage of the encoder to obtain a set of visual embeddings.
In detail, $\bm{V}_2$ is first projected into a hidden space by several convolution layers which convert its channel number from $C^v_2$ to $C^h$:
\begin{equation}
    \bm{V}^h = {\rm{Conv}}_{1 \times 1}(\delta({\rm{Conv}}_{3 \times 3}(\delta({\rm{Conv}}_{1 \times 1}(\bm{V}_2))))),
\end{equation}
where $\delta$ denotes the ReLU activation.
Then, we adjust the shape of $\bm{V}^h$ from $\mathbb{R}^{H_2 \times W_2 \times C^h}$ to $\mathbb{R}^{C^h \times H_2W_2}$ and apply a three-layer MLP on it to aggregate spatial information by regarding $H_2W_2$ as the input channel dimension for MLP:
\begin{equation}
    \bm{V}^e = {\rm{Reshape}}({\rm{MLP}}({\rm{Reshape}}(\bm{V}^h))),
\end{equation}
where the output of $\rm{MLP}$ has the shape of $\mathbb{R}^{C^h \times N}$, $\bm{V}^e \in \mathbb{R}^{N \times C^h}$ is a set of visual embeddings after another shape adjustment and $N$ is the number of visual embeddings.

\textbf{Audio prototype prompting.}
To enable more targeted audio information fetching of vision-derived queries in the subsequent Transformer decoder, we define a group of audio prototypes to prompt them with audio event categories present in the scene during the query generation.
In detail, we randomly initialize a set of learnable parameters as the audio prototypes $\bm{P} \in \mathbb{R}^{K \times C^h}$, where $K$ is the number of audio event categories, to prompt visual embeddings $\bm{V}^e$ with audio category priors by cross-attention:
\begin{equation}
    \Bar{\bm{V}}^e = \bm{V}^e + {\rm{Softmax}}(\frac{(\bm{V}^e\bm{W}^q_1)(\bm{P}\bm{W}^k_1)^{\rm{T}}}{\sqrt{C^h}})(\bm{P}\bm{W}^v_1).
\end{equation}
$\bm{W}^q_1$, $\bm{W}^k_1$ and $\bm{W}^v_1$ are projection weights.
For clarity, we omit other operations like $\rm{LN}$ and $\rm{FFN}$ and use $\Bar{\bm{V}}^e$ to denote the output of cross-attention.

Since the audio prototypes are randomly initialized, we also design a prototype-audio contrastive loss $\mathcal{L}_{\rm{pac}}$ to ensure that the prototypes can learn semantic information of different audio event categories from the audio features.
Concretely, let $\bm{A}_t \in \mathbb{R}^{S \times C^a}$ denotes the audio feature associated with the current frame and we perform linear projection followed by global average pooling on $\bm{A}_t$ to obtain $\Bar{\bm{A}}_t \in \mathbb{R}^{C^h}$.
An inner product between $\Bar{\bm{A}}_t$ and $\bm{P}$ yields a matching prediction $\bm{M} \in \mathbb{R}^{K}$ representing the likelihood of each audio category prototype present in the current audio.
The matching ground truth $\bm{M}^*$ indicating which audio event categories are actually present in the current audio can be obtained from the dataset annotation.
We implement $\mathcal{L}_{\rm{pac}}$ as a BCE loss between $\bm{M}$ and $\bm{M}^*$:
\begin{equation}
    \begin{split}
        \mathcal{L}_{\rm{bce}}(\bm{M}_k, \bm{M}^*_k) = &-\bm{M}^*_k\log(\bm{M}_k) \\
        &- (1 - \bm{M}^*_k)\log(1 - \bm{M}_k),
    \end{split}
\end{equation}
\begin{equation}
    \mathcal{L}_{\rm{pac}}(\bm{M}, \bm{M}^*) = \frac{1}{K}\sum_{k=1}^K\mathcal{L}_{\rm{bce}}(\bm{M}_k, \bm{M}^*_k).
\end{equation}
The $\mathcal{L}_{\rm{pac}}$ pulls the audio features closer to the present audio prototypes, while pushing the audio features further away from non-existent audio prototypes.

\textbf{Pixel context grouping.}
Inspired by GroupViT~\cite{xu2022groupvit}, to obtain more distinguishable vision-derived queries that can focus on different image regions, we adopt $\rm{Gumbel}$-$\rm{Softmax}$~\cite{jang2016categorical, maddison2016concrete} to group pixel context for each visual embedding with a hard and differentiable assignment:
\begin{equation}
    \bm{R} = {\rm{Softmax}}((\Bar{\bm{V}}^e\bm{W}^q_2)(\bm{V}^h\bm{W}^k_2)^{\rm{T}} + \bm{G}),
\end{equation}
\begin{equation}
    \Hat{\bm{R}} = {\rm{One\text{-}hot}}(\arg\max_{N}(\bm{R})) + \bm{R} - {\rm{sg}}(\bm{R}),
\end{equation}
where $\bm{W}^q_2$ and $\bm{W}^k_2$ are linear projection weights, $\bm{G} \in \mathbb{R}^{N \times H_2W_2}$ are i.i.d random samples drawn from the $\rm{Gumbel}(0, 1)$ distribution.
We apply $\arg\max$ on the $N$ dimension of $\bm{R} \in \mathbb{R}^{N \times H_2W_2}$ to find the index of the visual embedding with the highest similarity to each pixel feature as the assignment target.
$\rm{One\text{-}hot}$ converts the index for each pixel feature into a one-hot vector.
$\rm{sg}$ is the stop gradient operator.
The resulting $\Hat{\bm{R}} \in \mathbb{R}^{N \times H_2W_2}$ contains the hard assignments between visual embeddings and pixel features.
We use $\Hat{\bm{R}}$ to group the pixel context in image feature $\bm{V}^h$ into prototype-prompted visual embeddings $\Bar{\bm{V}}^e$:
\begin{equation}
    \bm{V}^q = \Bar{\bm{V}}^e + ({\rm{Norm}}(\Hat{\bm{R}})(\bm{V}^h\bm{W}^v_2))\bm{W}^o,
\end{equation}
where $\bm{W}^v_2$ and $\bm{W}^o$ are projection weights, $\rm{Norm}$ is normalization on the $H_2W_2$ dimension of $\Hat{\bm{R}}$ by dividing the sum of values, $\bm{V}^q \in \mathbb{R}^{N \times C^h}$ denotes the vision-derived queries generated by our PPQG module.

\subsection{Iterative Audio-Visual Transformer Decoder}
\label{sec:method_trans_dec}

In our designed iterative audio-visual Transformer decoder, vision-derived queries interact with the audio feature and multi-scale visual features iteratively through two kinds of Transformer blocks, each composed of a multi-head cross-attention (MHCA) layer, a multi-head self-attention (MHSA) layer, and an FFN layer.
In the audio information fetching block, each vision-derived query interacts with the audio feature of the current frame through MHCA, with $\bm{A}_t$ serving as the key and value.
MHCA allows each vision-derived query to obtain the corresponding sound information for the region it represents, enabling the determination of whether an object in that region is emitting sound and its audio category.
In the visual information enhancement blocks, each vision-derived query sequentially interacts with three different scales of visual features from the current frame, with $\bm{V}_5$, $\bm{V}_4$ and $\bm{V}_3$ serving as the key and value for MHCA respectively.
Here, we follow Mask2Former~\cite{Mask2Former} to use the segmentation mask predicted in the previous layer as the attention mask for the current layer, restricting the MHCA's attention to potential sounding image regions.
Through visual information enhancement, each vision-derived query captures finer-grained visual features, allowing for more accurate mask predictions.

We use $\mathcal{U} = \{\bm{A_t},\bm{V}_5, \bm{V}_4, \bm{V}_3\}$ as an interaction unit where one audio information fetching block and three visual information enhancement blocks are stacked in order.
We repeat $\mathcal{U}$ for $D$ times and append one audio block at the end to construct our iterative audio-visual Transformer decoder.
Through the above interaction, vision-derived queries gradually become aware of their corresponding sound information and fine-grained visual features, thus better distinguishing between different sounding objects and mitigating the perception ambiguity caused by mixed audio.

\subsection{Loss Functions}
\label{sec:method_loss_func}

Following Mask2Former~\cite{Mask2Former}, we use a classification head and a mask head to obtain category and segmentation predictions.
Our loss functions mainly consist of three parts, \textit{i.e.}, classification loss $\mathcal{L}_{\rm{cls}}$, mask loss $\mathcal{L}_{\rm{mask}}$, and our proposed prototype-audio contrastive loss $\mathcal{L}_{\rm{pac}}$.
The classification loss is namely CE loss.
Apart from BCE loss, the mask loss also contains Dice loss~\cite{Diceloss} to handle relatively small foreground areas in the images.
Therefore, the overall loss function of our method is formulated as:
\begin{equation}
    \mathcal{L} = \lambda_{\rm{cls}}\mathcal{L}_{\rm{cls}} + \lambda_{\rm{mask}}\mathcal{L}_{\rm{mask}} + \lambda_{\rm{pac}}\mathcal{L}_{\rm{pac}},
\end{equation}
where $\lambda_{\rm{cls}}$, $\lambda_{\rm{mask}}$ and $\lambda_{\rm{pac}}$ are loss coefficients.
\section{Experiments}
\label{sec:experiments}

\begin{table*}[t]
\setlength{\tabcolsep}{8pt}
\centering
\setlength{\arrayrulewidth}{0.6pt}
\resizebox{\linewidth}{!}{
\footnotesize
\begin{tabular}{l|c|c|c|cc|cc|cc}
\hline
\rowcolor[gray]{.9} & & & & \multicolumn{2}{c|}{\textbf{AVS-Semantic}} & \multicolumn{2}{c|}{\textbf{Single-Source}} & \multicolumn{2}{c}{\textbf{Multi-Source}} \\
\rowcolor[gray]{.9} & & & & \multicolumn{2}{c|}{\textbf{(AVSS)}} & \multicolumn{2}{c|}{\textbf{(S4)}} & \multicolumn{2}{c}{\textbf{(MS3)}} \\
\rowcolor[gray]{.9} \multirow{-3}{*}{\textbf{Method}} & \multirow{-3}{*}{\textbf{Reference}} & \multirow{-3}{*}{\textbf{Backbone}} & \multirow{-3}{*}{\textbf{Image Size}} & $\mathcal{M}_{\mathcal{J}}$ & $\mathcal{M}_{\mathcal{F}}$ & $\mathcal{M}_{\mathcal{J}}$ & $\mathcal{M}_{\mathcal{F}}$ & $\mathcal{M}_{\mathcal{J}}$ & $\mathcal{M}_{\mathcal{F}}$ \\
\hline\hline
\multirow{2}{*}{TPAVI~\cite{TPAVI,TPAVI_J}} & \multirow{2}{*}{\pub{ECCV'22}} & ResNet-50 & \multirow{2}{*}{224$\times$224} & - & - & 72.8 & 84.8 & 46.9 & 57.8 \\
&  & PVT-v2 & & 29.8 & 35.2 & 78.7 & 87.9 & 54.0 & 64.5 \\
\hline
\multirow{2}{*}{AQFormer~\cite{AQFormer}} & \multirow{2}{*}{\pub{IJCAI'23}} & ResNet-50 & \multirow{2}{*}{224$\times$224} & - & - & 77.0 & 86.4 & 55.7 & 66.9 \\
&  & PVT-v2 & & - & - & 81.6 & 89.4 & 61.1 & \underline{72.1} \\
\hline
\multirow{2}{*}{ECMVAE~\cite{ECMVAE}} & \multirow{2}{*}{\pub{ICCV'23}} & ResNet-50 & \multirow{2}{*}{224$\times$224} & - & - & 76.3 & 86.5 & 48.7 & 60.7 \\
&  & PVT-v2 & & - & - & 81.7 & 90.1 & 57.8 & 70.8 \\
\hline
\multirow{2}{*}{CATR~\cite{CATR}} & \multirow{2}{*}{\pub{ACMMM'23}} & ResNet-50 & \multirow{2}{*}{224$\times$224} & - & - & 74.8 & 86.6 & 52.8 & 65.3 \\
 &  & PVT-v2 & & 32.8 & 38.5 & 81.4 & 89.6 & 59.0 & 70.0 \\
 \hline
\multirow{2}{*}{AVSC~\cite{AVSC}} & \multirow{2}{*}{\pub{ACMMM'23}} & ResNet-50 & \multirow{2}{*}{224$\times$224} & - & - & 77.0 & 85.2 & 49.6 & 61.5 \\ 
 &  & PVT-v2 & & - & - & 80.6 & 88.2 & 58.2 & 65.1 \\
 \hline
\multirow{3}{*}{BAVS~\cite{BAVS}} & \multirow{3}{*}{\pub{TMM'24}} & ResNet-50 & \multirow{3}{*}{224$\times$224} & 24.7 & 29.6 & 78.0 & 85.3 & 50.2 & 62.4 \\
&  & PVT-v2 & & 32.6 & 36.4 & 82.0 & 88.6 & 58.6 & 65.5 \\
&  & Swin-B & & \underline{33.6} & \underline{37.5} & \underline{82.7} & \underline{89.8} & \underline{59.6} & \underline{65.9} \\
\hline
\multirow{2}{*}{AVSBG~\cite{AVSBG}} & \multirow{2}{*}{\pub{AAAI'24}} & ResNet-50 & \multirow{2}{*}{224$\times$224} & - & - & 74.1 & 85.4 & 45.0 & 56.8 \\
&  & PVT-v2 & & - & - & 81.7 & 90.4 & 55.1 & 66.8 \\
\hline
GAVS~\cite{GAVS} & \pub{AAAI'24} & ViT-B & 224$\times$224 & - & - & 80.1 & 90.2 & 63.7 & 77.4 \\
\hline
\multirow{2}{*}{AVSF~\cite{AVSegFormer}} & \multirow{2}{*}{\pub{AAAI'24}} & ResNet-50 & \multirow{2}{*}{224$\times$224} & 24.9 & 29.3 & 76.5 & 85.9 & 49.5 & 62.8 \\ 
&  & PVT-v2 & & 36.7 & 42.0 & 82.1 & 89.9 & 58.4 & 69.3 \\
\hline
\multirow{2}{*}{UFE~\cite{UFE}} & \multirow{2}{*}{\pub{CVPR'24}} & ResNet-50 & \multirow{2}{*}{224$\times$224} & - & - & 79.0 & 87.5 & 55.9 & 64.5 \\
&  & PVT-v2 & & - & - & 83.2 & 90.4 & \textbf{62.0} & 70.9 \\
\hline
CAVP~\cite{CAVP} & \pub{CVPR'24} & ResNet-50 & 224$\times$224 & 30.4 & 35.3 & 78.8 & 88.9 & 55.8 & 67.1 \\
\hline
\multirow{2}{*}{COMBO~\cite{COMBO}} & \multirow{2}{*}{\pub{CVPR'24}} & ResNet-50 & \multirow{2}{*}{224$\times$224} & 33.3 & 37.3 & \underline{81.7} & 90.1 & 54.5 & 66.6 \\
&  & PVT-v2 & & \underline{42.1} & 46.1 & \underline{84.7} & \underline{91.9} & 59.2 & 71.2 \\
\hline
CPM~\cite{CPM} & \pub{ECCV'24} & ResNet-50 & 224$\times$224 & \underline{34.5} & \underline{39.6} & 81.4 & \underline{90.5} & \underline{59.8} & \underline{71.0} \\
\hline
TeSO~\cite{TeSO} & \pub{ECCV'24} & Swin-B & 384$\times$384 & 39.0 & 45.1 & \underline{83.3} & \underline{93.3} & 66.0 & 80.1 \\
\hline
\multirow{2}{*}{SelM~\cite{SelM}} & \multirow{2}{*}{\pub{ACMMM'24}} & ResNet-50 & \multirow{2}{*}{224$\times$224} & 31.9 & 37.2 & 76.6 & 86.2 & 54.5 & 65.6 \\ 
&  & PVT-v2 & & 41.3 & \underline{46.9} & 83.5 & 91.2 & 60.3 & 71.3 \\
\hline
AVSBias~\cite{AVSBias} & \pub{ACMMM'24} & Swin-B & 384$\times$384 & \underline{44.4} & \underline{49.9} & \underline{83.3} & 93.0 & \underline{67.2} & \underline{80.8} \\
\hline
\rowcolor{blue!8} & & ResNet-50 & 224$\times$224  & \textbf{37.5} & \textbf{42.2} & \textbf{81.8} & \textbf{90.6} & \textbf{61.9} & \textbf{74.7} \\
\rowcolor{blue!8} & & PVT-v2 & 224$\times$224 & \textbf{44.7} & \textbf{49.5} & \textbf{84.8} & \textbf{92.1} & \textbf{62.0} & \textbf{75.0} \\ 
\rowcolor{blue!8}\multirow{-2}{*}{\textbf{VCT~(Ours)}} & \multirow{-2}{*}{-} & Swin-B & 224$\times$224  & \textbf{47.9} & \textbf{52.9} & \textbf{84.7} & \textbf{92.3} & \textbf{67.5} & \textbf{79.3} \\
\rowcolor{blue!8} & & Swin-B & 384$\times$384 & \textbf{51.2} & \textbf{55.5} & \textbf{86.2} & \textbf{93.4} & \textbf{67.6} & \textbf{81.4} \\
\hline
\end{tabular}}
\caption{Comparison with state-of-the-art methods on three subsets of the AVSBench dataset. Our method obtains significant performance gains on the most challenging AVSS subset. Best results in \textbf{bold} and 2$^{\rm{nd}}$ best \underline{underlined} for different backbones and image sizes.}
\label{tab:sota_avsbench}
\end{table*}

\subsection{Dataset and Evaluation Metrics}

We conduct experiments on the AVSBench dataset~\cite{TPAVI, TPAVI_J} to train and evaluate our models.
The AVSBench dataset is organized into three subsets: the Single-source subset (S4), the Multi-sources subset (MS3), and the Semantic-labels subset (AVSS).
% The S4 subset includes 4,932 videos across 23 distinct categories, with each video containing only one sound-producing object.
% The MS3 subset comprises 424 videos featuring multiple audio sources.
Built on the previous two subsets, the AVSS subset substantially extends the dataset in both quantity and semantic labels, which brings more challenges by requiring models not only to produce segmentation masks but also to categorize each sounding object.
% Both S4 and MS3 subsets focus on binary foreground object segmentation.
% Built on the two subsets, the AVSS subset substantially extends the dataset to 12,356 videos in 70 categories and brings more challenges by incorporating semantic labels, which require models not only to produce segmentation masks but also categorize each sounding object.

Regarding evaluation metrics, we follow prior works~\cite{TPAVI, TPAVI_J, COMBO} to adopt $\mathcal{M}_{\mathcal{J}}$ and $\mathcal{M}_{\mathcal{F}}$ for performance measurement.
$\mathcal{M}_{\mathcal{J}}$ is the Jaccard index which computes the Intersection over Union (IoU) between predicted segmentation and ground truth.
On the AVSS subset, $\mathcal{M}_{\mathcal{J}}$ is the mean IoU averaged over all categories.
$\mathcal{M}_{\mathcal{F}}$ denotes the F-score which combines the precision and recall of the segmentation results.
Its formulation is $\mathcal{M}_{\mathcal{F}} = \frac{(1 + \beta^2) \cdot \text{Precision} \cdot \text{Recall}}{\beta^2 \cdot \text{Precision} + \text{Recall}}$ where $\beta^2 = 0.3$ indicating more emphasis on the recall.

\subsection{Implementation Details}
Our model is implemented with PyTorch.
For a fair comparison with previous methods~\cite{COMBO, AAVS, TeSO}, we adopt ResNet-50~\cite{ResNet} and Pyramid Vision Transformer (PVT-v2)~\cite{PVT-v2} pretrained on ImageNet~\cite{ImageNet} and Swin Transformer (Swin-Base)~\cite{liu2021swin} pretrained on ImageNet-21K~\cite{imagenet21k}  as visual encoders.
The audio encoder is VGGish~\cite{VGGish} pretrained on YouTube-8M~\cite{abu2016youtube}.
Following COMBO~\cite{COMBO}, we also adopt Semantic-SAM~\cite{li2024segment} to generate colored class-agnostic Maskige as segmentation prior knowledge and fuse its feature with the original visual feature in the same way as COMBO.
The values of $C^h$ and $S$ are $256$ and $24$.
The number of vision-derived queries $N = 100$.
$D = 2$ in the iterative audio-visual Transformer decoder.
$\lambda_{\rm{cls}}$, $\lambda_{\rm{mask}}$, and $\lambda_{\rm{pac}}$ are set to $2$, $5$, and $1$.
The AdamW optimizer is used with the learning rate of $1e^{-4}$.
We train our model with a batch size of $16$ for ResNet-50 and $8$ for PVT-v2 and Swin-B.
The training iterations are $45$K, $40$K, and $45$K for S4, MS3, and AVSS subsets respectively.

\subsection{Comparison with State-of-the-art Methods}

In Table~\ref{tab:sota_avsbench}, we show the performance comparison between our method and previous state-of-the-art methods on the three subsets of the AVSBench dataset.
Our method outperforms the previous best-performing approaches across three subsets with different visual backbones and image sizes, demonstrating its ability to effectively differentiate multiple sound sources, accurately delineate the corresponding object shapes, and recognize their categories.
In particular, on the most challenging AVSS subset, our method achieves even better $\mathcal{M}_{\mathcal{J}}$ performance with a PVT-v2 backbone and a $224\times224$ image size compared to AVSBias~\cite{AVSBias} using a Swin-B backbone and a $384\times384$ image size.
This also indicates that our new architecture, which integrates vision-derived queries and the iterative audio-visual Transformer, demonstrates a clear advantage over the previous architectures that rely on audio-derived queries to interact with visual information.
Moreover, on the S4 subset where performance is relatively plateaued, our method achieves 86.2 $\mathcal{M}_{\mathcal{J}}$, indicating that it is better at handling some of the hard cases compared to previous methods.

\subsection{Ablation Studies}

We conduct ablation studies on the AVSS subset with ResNet-50 as the visual backbone to verify the effectiveness of different designs in our method.

\begin{table}[!htbp]
\centering
\setlength{\arrayrulewidth}{0.4pt}
\resizebox{\linewidth}{!}{
\tiny
\begin{tabular}{c|l|c|c}
\hline
\rowcolor[gray]{.9} \textbf{Framework} & \textbf{Queries} & $\mathcal{M}_{\mathcal{J}}$ & $\mathcal{M}_{\mathcal{F}}$ \\
\hline\hline
\textbf{ACT} & Audio-Derived Queries & 33.2 & 37.0 \\
\hline
\multirow{4}{*}{\textbf{VCT}} & Naive Vision Queries & 35.2 & 39.3 \\
& \textit{w/} Cross-Attention & 35.8 & 39.8 \\
& \textit{w/} Group-Attention & 36.3 & 40.5 \\
& \textbf{\textit{w/} Audio Prototypes} & \textbf{37.5} & \textbf{42.2} \\
\hline
\end{tabular}}
\caption{Ablation study of generating vision-derived queries in our PPQG module on the AVSS subset.}
\label{tab:ablation_component}
\end{table}

\textbf{Component analysis.}
We include the ablation study results of generating vision-derived queries in Table~\ref{tab:ablation_component} to show the influence of different operations in our proposed PPQG module.
``ACT'' denotes the baseline model with audio-derived queries where audio features largely enhance the initial learnable queries for later audio-visual interaction.
In our VCT framework, ``Naive Vision Queries'' represents that the vision-derived queries are generated only by the first step of visual embedding aggregation in the PPQG module.
It yields better performance than audio-derived queries, which indicates that using visual features as the basis for locating sound-emitting objects, compared to introducing entangled audio information, enables better visual-audio correspondence during subsequent decoder interactions.
``\textit{w/} Cross-Attention'' denotes that we use cross-attention with normal softmax to incorporate image features into naive vision-derived queries, while ``\textit{w/} Group-Attention'' means replacing cross-attention with group-attention equipped by Gumbel-softmax to group pixel contexts for vision-derived queries using hard assignment.
The performance improvements brought by both attention mechanisms show that incorporating additional visual region information helps to obtain more distinguishable vision-derived queries.
The last row demonstrates that providing audio category priors to vision-derived queries enables them to more effectively fetch the corresponding audio information in the subsequent iterative audio-visual Transformer decoder.

\begin{table}[!htbp]
\centering
\setlength{\arrayrulewidth}{0.4pt}
\resizebox{\linewidth}{!}{
\tiny
\begin{tabular}{l|c|c}
\hline
\rowcolor[gray]{.9}\textbf{Method} & $\mathcal{M}_{\mathcal{J}}$ & $\mathcal{M}_{\mathcal{F}}$ \\
\hline\hline
\textit{w/o} Audio Prototypes & 36.3 & 40.5 \\
\hline
\textit{w/} Audio Prototypes, \textit{w/o} loss & 36.3 & 40.4 \\
\textit{w/} Audio Prototypes, \textit{w/} visual loss & 36.5 & 40.8 \\
\textbf{\textit{w/} Audio Prototypes, \textit{w/} PAC loss} & \textbf{37.5} & \textbf{42.2} \\
\hline
\end{tabular}}
\caption{Ablation study of audio prototype prompting in PPQG.}
\label{tab:ablation_prototypes}
\end{table}

\begin{figure*}[!t]
    \centering
    \includegraphics[width=\linewidth]{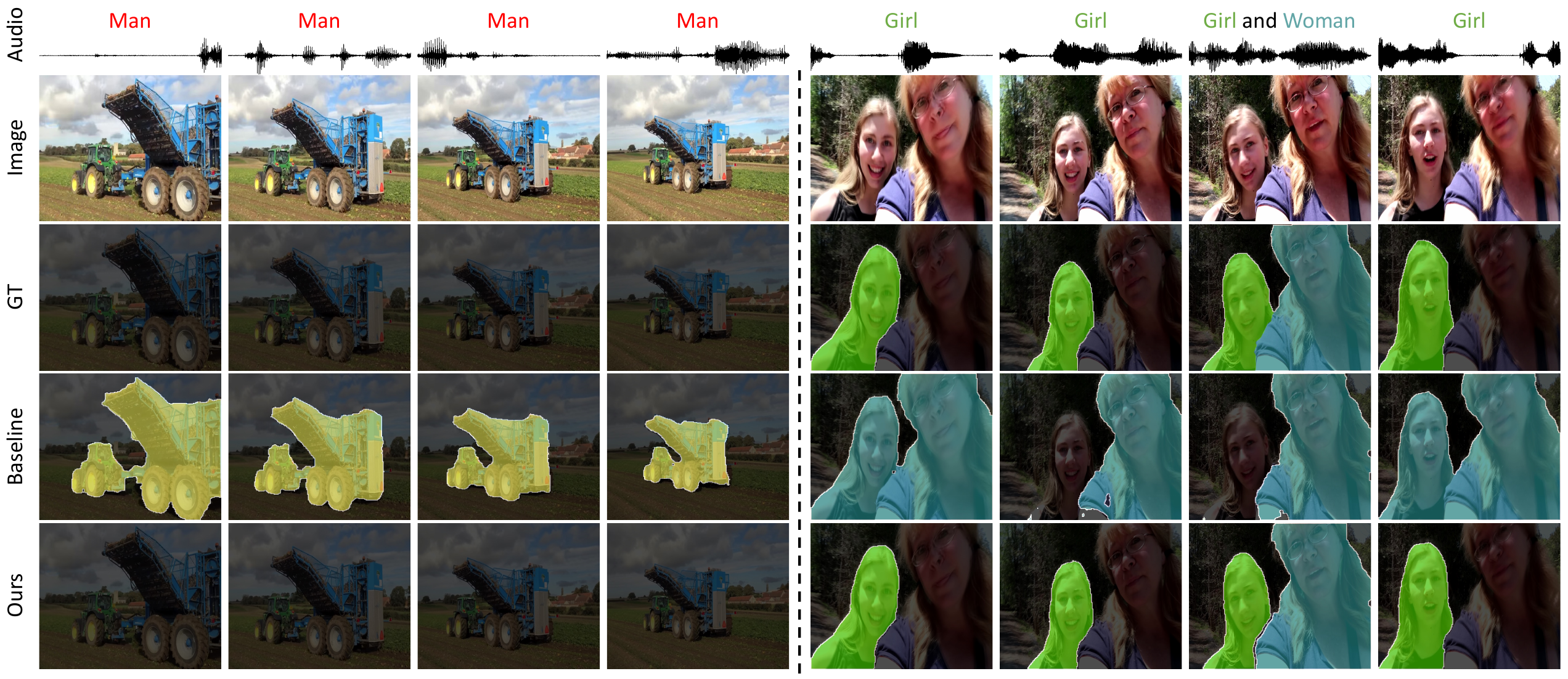}
    \caption{Qualitative comparison between our full model and the ACT baseline. Existing sounds correspond to masks of the same colors.}
    \label{fig:seg_results}
\end{figure*}

\begin{figure}[!htbp]
    \centering
    \includegraphics[width=\linewidth]{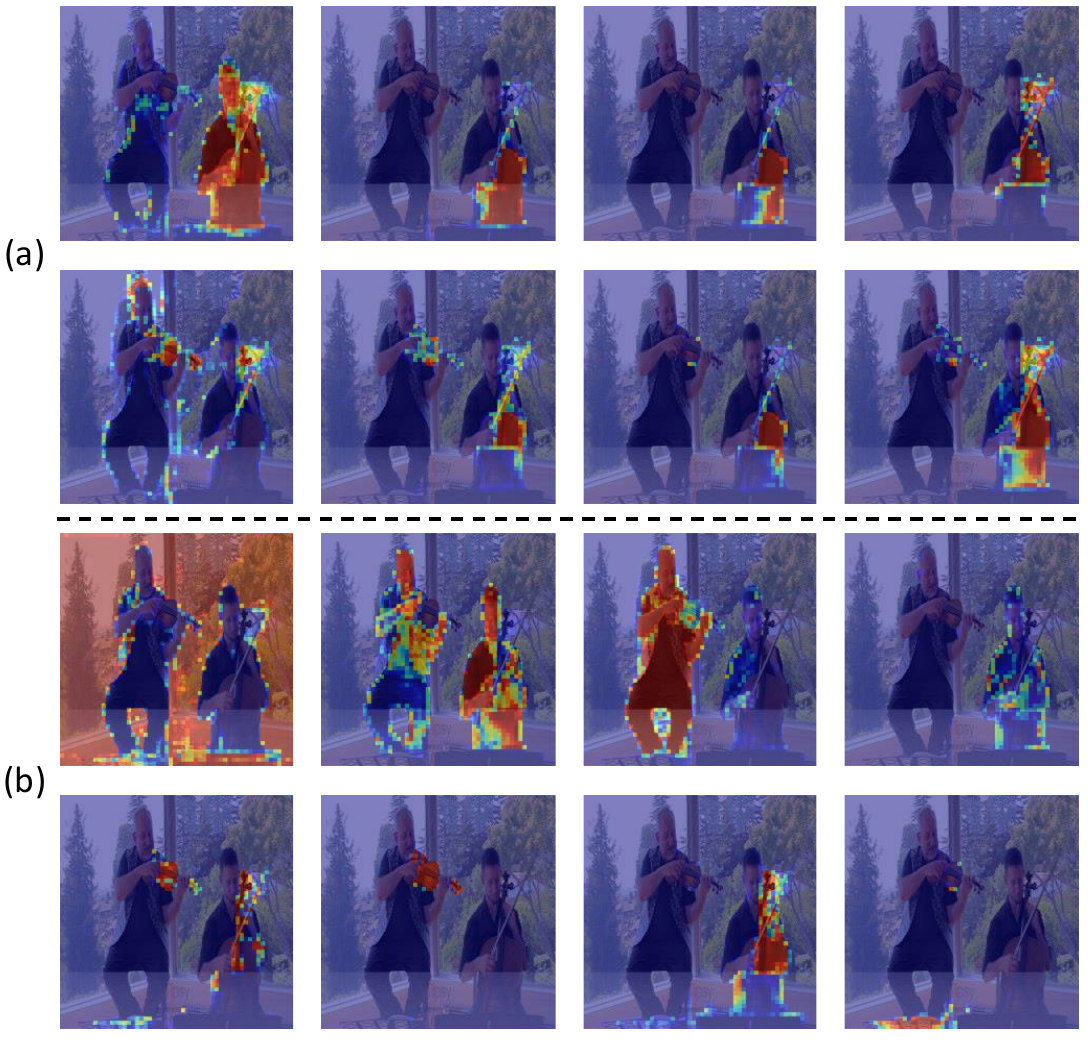}
    \caption{Visualization of logit maps from different types of queries. (a) Audio-derived queries. (b) Vision-derived queries.}
    \label{fig:logits_vis}
\end{figure}

\textbf{Audio prototype prompting.}
Table~\ref{tab:ablation_prototypes} summarizes the ablation results of audio prototype prompting in our PPQG module.
By comparing the first and second rows, it can be observed that without loss constraints, the randomly initialized audio prototypes fail to learn meaningful audio event category priors, resulting in no positive impact on performance.
In the third row, ``\textit{w/} visual loss'' denotes that we compute contrastive loss between audio prototypes and positive vision-derived queries that matched with the ground truth, bringing a small performance gain.
In contrast, introducing our designed PAC loss in the last row results in an obvious performance improvement, suggesting that audio prototypes, through contrastive learning with audio features, can learn more explicit and clear audio event category priors than those learned from positive vision queries.

\begin{table}[!htbp]
\centering
\setlength{\arrayrulewidth}{0.4pt}
\resizebox{\linewidth}{!}{
\tiny
\begin{tabular}{c|l|c|c}
\hline
\rowcolor[gray]{.9} & \textbf{Audio Feature} & $\mathcal{M}_{\mathcal{J}}$ & $\mathcal{M}_{\mathcal{F}}$ \\
\hline\hline
\multirow{3}{*}{\textbf{Visual Feature}} & Multiply & 33.9 & 37.9 \\
& Concatenation & 35.3 & 39.3 \\
& Addition & 36.3 & 40.5 \\
\hline
\textbf{Audio Proto} & Replace & 36.6 & 40.8 \\
\hline
\textbf{VCT} & \textbf{Full Model} & \textbf{37.5} & \textbf{42.2} \\
\hline
\end{tabular}}
\caption{Ablation study of audio feature incorporation.}
\label{tab:ablation_audio_feat}
\end{table}

\textbf{Audio feature incorporation.}
We also explore different approaches to incorporate audio features into our VCT framework and present the results in Table~\ref{tab:ablation_audio_feat}.
The first three rows show how we fuse audio features with visual features using multiplication, concatenation, and addition, which are then used as the key and value in the cross-attention of the Transformer decoder.
The fourth row denotes that we replace the audio prototypes in PPQG with the audio feature and the PAC loss is hence removed.
Our full VCT obtains the best performance compared to these approaches, indicating that pre-fetching audio features during the generation of vision-derived queries or fetching fused multimodal features in the Transformer decoder can lead to queries learning inaccurate audio-visual correspondence.
Additional ablation results are included in the supplementary materials.

\subsection{Qualitative Analysis}

Figure~\ref{fig:seg_results} presents the qualitative comparison between our full model and the ACT baseline.
In the left example, the input audio is the voiceover narration.
Compared to the baseline, our full model is able to recognize the off-screen background audio rather than biasing its focus on the main objects in the video, demonstrating stronger robustness.
In the right example, the baseline confuses between the girl and woman, while our full model correctly distinguishes between them and identifies the actual sound-emitting source in the video.
This demonstrates that the vision-derived queries have a stronger ability to differentiate between different regions in the image and can fetch the corresponding information from the audio.

In addition, we also visualize the logit maps predicted by audio-derived and vision-derived queries within the Transformer decoder.
For each type of 100 queries, we uniformly select one every five queries, remove those with blank logit maps, and then randomly retain 8 queries for display.
As shown in Figure~\ref{fig:logits_vis}(a), both instruments are producing sound in this example, but most of the audio-derived queries focus on the person on the right and the cello, indicating that these queries are biased towards certain sound-emitting objects due to the influence of entangled audio.
In contrast, our vision-derived queries Figure~\ref{fig:logits_vis}(b) are able to focus on diverse visual regions in the image, demonstrating that vision-derived queries are more distinguishable, which forms the foundation for superior performance.

\section{Conclusion}
\label{sec:conclusion}

Given an audio clip, the goal of audio-visual segmentation (AVS) is to segment the sounding objects in the video.
In this work, we proposed a new Vision-Centric Transformer (VCT) framework to overcome the limitations of audio-centric Transformers adopted by prevailing AVS methods.
Concretely, our VCT framework utilizes vision-derived queries to iteratively extract both corresponding audio and fine-grained visual information, which aids in better distinguishing between different sound-emitting objects, eliminating the perception ambiguity caused by mixed audio and yielding accurate object segmentation.
Additionally, we introduce a Prototype Prompted Query Generation (PPQG) module within the VCT framework, which generates vision-derived queries that contain both rich visual details and audio semantics through audio prototype prompting and pixel context grouping, facilitating later information aggregation.
Extensive experiments on three subsets of the AVSBench dataset demonstrate that our method achieves new state-of-the-art performances.

\section{Acknowledgments}
This research is supported in part by National Key R\&D Program of China (2022ZD0115502), National Natural Science Foundation of China (No. 62461160308, U23B2010), ``Pioneer'' and ``Leading Goose'' R\&D Program of Zhejiang (No. 2024C01161).

{
    \small
    \bibliographystyle{ieeenat_fullname}
    \bibliography{main}
}


\end{document}